\documentclass[a4paper,twoside]{article}

\usepackage{epsfig}
\usepackage{subcaption}
\usepackage{calc}
\usepackage{amssymb}
\usepackage{amstext}
\usepackage{amsmath}
\usepackage{amsthm}
\usepackage{multicol}
\usepackage{pslatex}
\usepackage{apalike}
\usepackage{algorithm2e}
\usepackage{booktabs}
\usepackage{graphicx}
\usepackage{url}
\usepackage[bottom]{footmisc}
\usepackage{SCITEPRESS}     

\newcommand{\DB}{\Delta\mathcal{B}}
\newcommand{\sys}{\textsc{VerifyAgent}}
\newcommand{\citep}[1]{\cite{#1}}

\begin{document}

\title{Beyond Self-Critique: Epistemic Convergence Control for Multi-Agent Research Verification}

\author{\authorname{Chrisantus Eze}
\affiliation{Oklahoma State University}
\email{chrisantus.eze@okstate.edu}
}


\abstract{Multi-agent systems for scientific claim verification increasingly rely on role-structured LLM debate to pressure-test hypotheses. However, when debating agents are instantiated from the same underlying base model, they share identical latent representations, making self-critique bias structurally inevitable. Static fixed-iteration protocols compound this by either halting prematurely before evidence stabilizes or wasting inference compute after convergence. We present \sys{}, a multi-agent research verification framework that addresses both failure modes. First, we introduce an asymmetric role architecture comprising an Advocate, Skeptic, and Arbitrator with opposing epistemic directives, producing significantly lower role-embedding similarity (0.852 vs.\ 0.918 for symmetric Multi-Agent Debate [MAD], $p = 5.82 \times 10^{-8}$) and surfacing more contested claims (22.0\% vs.\ 15.4\%). Second, we propose the Belief-Delta ($\DB$) convergence criterion, an evidence-grounded stopping rule that eliminates over-iteration (0.0\% vs.\ 62.5\% for Fixed-$3$) while cutting LLM calls by 2.4$\times$. Third, each supported claim is anchored to a verbatim literature snippet audited by an independent NLI classifier, achieving a 21.7\% hallucination rate (comparable to 25.9\% for a Self-Refine critic, $p = 0.61$) while establishing auditable evidence grounding. Across 200 runs on 40 research hypotheses evaluated with DeepSeek-R1-32B, \sys{} balances compute efficiency with rigorous epistemic verification.}

\makeatletter
\g@addto@macro\@abstract{%
  \par\vspace{0.35em}%
  \begin{center}%
    \includegraphics[width=0.98\textwidth]{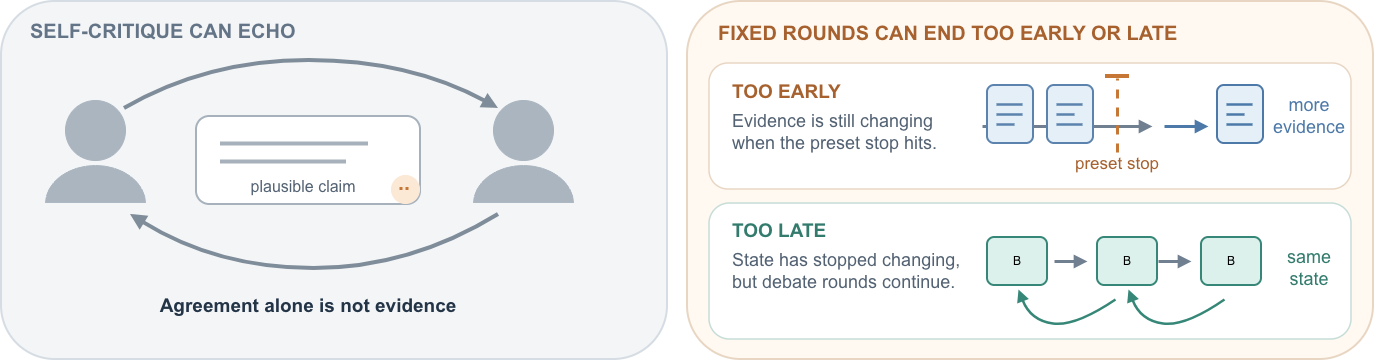}\par
    \vspace{-0.35em}%
    \captionof{figure}{Motivation for \sys{}. Self-critique can echo a plausible claim without independent evidence; fixed-round debate can stop while evidence is changing or continue after it stabilizes. These are conceptual design risks, not experimental findings.}\label{fig:motivation}%
  \end{center}%
  \vspace{0.4em}%
}
\makeatother

\onecolumn \maketitle \normalsize \setcounter{footnote}{0} \vfill

\section{\uppercase{Introduction}}
\label{sec:intro}

\noindent Figure~\ref{fig:motivation} illustrates the two design risks motivating this work: self-critique can echo claims without independent evidence, and fixed-round debate can mismatch the pace of evidence change.

\noindent The pursuit of automated scientific discovery has rapidly progressed from conceptual vision to practical reality~\citep{gottweis2025,lu2024aiscientist,schmidgall2025}. Large language model (LLM) agents now independently review literature, formulate candidate hypotheses, write experimental protocols, and draft academic manuscripts. Yet the epistemic reliability of such systems hinges on a critical question: \emph{how can an autonomous system verify whether a research claim is substantiated by published empirical evidence rather than confabulated through plausible-sounding text generation?} In scientific domains characterized by conflicting findings and nuanced methodological qualifications, separating empirically supported claims from contested assertions is non-trivial~\citep{hope2023scimon}.

While chain-of-thought prompting~\citep{wei2022chain} and reinforcement-learning-tuned reasoning models~\citep{deepseekai2025r1} exhibit strong single-turn deduction, foundational studies demonstrate that LLMs \emph{cannot reliably self-correct} their own reasoning errors without external ground-truth feedback~\citep{huang2023cannot,stechly2024selfcorrection}. When an agent critiques its own generation (as in Self-Refine~\citep{madaan2023} or unconstrained verbal reflection~\citep{shinn2023reflexion}), it frequently falls into self-confirmation bias, ratifying flawed premises rather than exposing contradictions~\citep{pan2023}.

Multi-agent debate (MAD) frameworks~\citep{liang2023,du2023,chan2023chateval} seek to overcome single-agent bias by distributing reasoning across debating personas. However, existing multi-agent verifiers share a fundamental structural limitation: all debating agents are typically instantiated from the same underlying base LLM. When an Advocate and a Skeptic evaluate a hypothesis using identical model weights, their arguments are sampled from a shared representation space. This is not genuine debate so much as \emph{self-critique with theatrics}, a failure mode wherein symmetric agents converge toward mutual sycophancy or cosmetic agreement~\citep{perez2022discovering,sharma2023sycophancy}. Prompt differentiation cannot change the underlying weights, but it can enforce divergent behavioral mandates and monitored representation divergence to prevent agents from defaulting to consensus.

A second challenge is \emph{when to terminate} the verification loop. Fixed-iteration schedules either cut off prematurely before literature evidence has been synthesized, or waste costly inference compute by running long after the epistemic state has stabilized. Self-report stopping, where the model declares its own termination based on verbalized certainty, suffers from well-documented confidence miscalibration~\citep{kadavath2022language}.

To address these challenges, we present \sys{}, an agentic verification system built on two core design principles: (1)~structurally decouple the \emph{evaluator} from the \emph{generator} via divergent epistemic mandates and real-time representation monitoring, so that verification cannot ratify its own priors; and (2)~define convergence as a measurable, artifact-grounded stabilization of the belief state rather than model self-reported confidence.

\smallskip\noindent\textbf{Contributions.}
\begin{enumerate}
  \item \textbf{Asymmetric Role Architecture:} We formalize an Advocate/Skeptic/Arbitrator interaction protocol with opposing epistemic incentives and continuous embedding divergence monitoring. Structural asymmetry achieves significantly lower role-embedding similarity ($0.852$ vs.\ $0.918$, $p = 5.82 \times 10^{-8}$) and higher contested claim surfacing (22.0\% vs.\ 15.4\%) compared to symmetric MAD.
  \item \textbf{Belief-Delta ($\DB$) Convergence Criterion:} We formalize an artifact-grounded stopping metric based on the transition rate in the structured belief state. The $\DB$ criterion eliminates over-iteration entirely (0.0\% vs.\ 62.5\%) while reducing LLM calls by 2.4$\times$ relative to Fixed-$3$.
  \item \textbf{Source-Grounded Evidence Links:} Every supported claim is tied to a verbatim snippet from retrieved literature and audited by an independent cross-encoder NLI model, providing structural auditability and a hallucination rate of 21.7\%.
\end{enumerate}

\begin{figure*}[!t]
  \centering
  \includegraphics[width=1.0\textwidth]{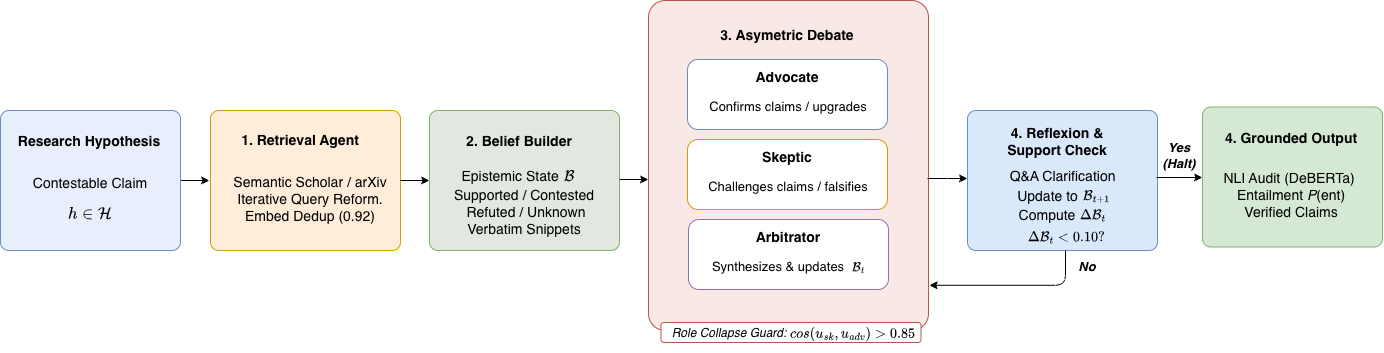}
  \caption{\sys{} Architecture: Candidate hypotheses are mapped to targeted queries by the Retrieval Agent. The Belief Builder compiles retrieved literature into an initial epistemic state $\mathcal{B}_0$. The Asymmetric Pressure Tester engages Advocate, Skeptic, and Arbitrator roles with continuous role-collapse monitoring. The Reflexion Node incorporates clarifications into $\mathcal{B}_{t+1}$ and evaluates the $\DB_t$ convergence criterion: iterating if $\DB_t \ge 0.10$, or halting to pass grounded claims to the NLI audit store.}
  \label{fig:architecture}
\end{figure*}

\section{\uppercase{Related Work}}
\label{sec:related}

\paragraph{Autonomous Scientific Agents.}
Agent Laboratory~\citep{schmidgall2025} coordinates autonomous LLM agents across literature review, experimentation, and report writing. The AI co-scientist~\citep{gottweis2025} employs specialist agent tournaments to refine biomedical hypotheses, and the AI Scientist~\citep{lu2024aiscientist} automates ML ideation and manuscript generation. SciMon~\citep{hope2023scimon} explores literature-grounded hypothesis generation. \sys{} occupies a distinct niche: rather than generating hypotheses, it provides an auditable \emph{epistemic verification pipeline} that pressure-tests candidate claims against published literature.

\paragraph{Multi-Agent Collaboration and Debate.}
Frameworks such as CAMEL~\citep{li2023camel} and AutoGen~\citep{wu2023autogen} study structured communication between agents. Liang et al.~\cite{liang2023} and Du et al.~\cite{du2023} report gains from multi-agent debate, while ChatEval~\citep{chan2023chateval} applies debate to evaluation. Prior systems also explore role-differentiated prompts, but generally do not measure role overlap with an embedding-based guard. Our experiment tests that measurement and prompt combination; it does not show that other debate systems necessarily collapse to consensus.

Several related results qualify the premise that shared model weights are the central obstacle. ReConcile uses heterogeneous LLMs and reports that model diversity contributes to reasoning gains, making heterogeneous panels an important untested alternative to prompt asymmetry~\citep{chen2023reconcile}. A broad comparison of debate and non-debate prompting strategies finds that debate does not reliably outperform simpler strategies without careful configuration~\citep{smit2023mad}. Khan et al. study an information-asymmetric debate in which expert debaters help a weaker judge; this supports debate as a possible oversight mechanism, while differing from our single-model, literature-grounded verification setting~\citep{khan2024persuasive}. More recently, Hu et al. propose adaptive debate stopping for LLM judges using distributional stability of consensus estimates~\citep{hu2025stability}. Our Belief-Delta rule instead tracks changes in a structured evidence-linked claim state. We do not compare against heterogeneous models or these alternative stopping methods.

\paragraph{Iterative Refinement and Self-Correction Limits.}
Self-Refine~\citep{madaan2023}, Reflexion~\citep{shinn2023reflexion}, and CRITIC~\citep{gou2024critic} enable multi-turn self-revision, but Huang et al.~\cite{huang2023cannot} and Stechly et al.~\cite{stechly2024selfcorrection} show that LLMs cannot self-correct without external ground truth; self-feedback often degrades performance. \sys{} seeks to reduce unsupported updates by tying belief entries to retrieved literature and auditing evidence links with an independent verifier.

\paragraph{Retrieval Grounding and Factuality Auditing.}
RAG~\citep{lewis2020retrieval} grounds generation in external documents, but retrieval alone does not guarantee factual consistency. FActScore~\citep{min2023factscore} and multi-agent hallucination studies~\citep{wei2024measuring} emphasize atomic claim verification. \sys{} combines atomic claim decomposition with mandatory verbatim snippet binding, evaluated via cross-encoder NLI~\citep{he2020deberta} and bi-encoder embeddings~\citep{reimers2019sentencebert}.

\section{\uppercase{System Architecture}}
\label{sec:architecture}

\noindent \sys{} is a stateful multi-agent pipeline implemented in LangGraph~\citep{langgraph2024}, consisting of five components illustrated in Figure~\ref{fig:architecture}.

\subsection{Retrieval Agent}
Given a candidate research hypothesis $h$, the Retrieval Agent iteratively queries Semantic Scholar and arXiv. Queries are reformulated dynamically when returned results are sparse or semantically redundant (measured via embedding cosine similarity). Retrieved papers are deduplicated at a cosine similarity threshold of 0.92, ranked by academic citation count, and passed to the Belief Builder.

\subsection{Epistemic Belief State}
The Belief Builder constructs a structured epistemic state $\mathcal{B}$, represented as a set of tuples $\langle \text{claim}, \text{status}, \text{evidence}\rangle$, where status takes one of four discrete values:
\begin{equation}
  \text{status} \in \left\{\begin{aligned}
    &\texttt{supported}, \texttt{contested}, \\
    &\texttt{refuted}, \texttt{unknown}
  \end{aligned}\right\}
\end{equation}
Each evidence entry is an \texttt{EvidenceLink} containing the source paper identifier, the atomic claim text, a verbatim 120-character \texttt{source\_snippet} excerpted from the abstract, and the stance. Belief states with empty snippets for \texttt{supported} claims are rejected and regenerated (up to three retries), ensuring every affirmative claim has a verifiable textual anchor.

\subsection{Asymmetric Pressure Tester}
Three structurally differentiated roles interact to pressure-test the candidate belief state:
\begin{itemize}
  \item \textbf{Advocate:} Solicits affirmative evidence from the retrieved literature, proposing upgrades from \texttt{unknown} to \texttt{supported}.
  \item \textbf{Skeptic:} Actively challenges \texttt{supported} entries, searching for methodological caveats, disconfirming empirical results, or over-generalizations.
  \item \textbf{Arbitrator:} Synthesizes opposing arguments and updates the status of each claim in $\mathcal{B}$.
\end{itemize}
Advocate and Skeptic are prompted with opposing epistemic directives and executed independently in isolated inference passes without mutual context, reducing direct co-generation of rationales.

A \emph{role-collapse guard} monitors role-embedding similarity as an operational proxy for representational divergence. Let $C_{\text{sk}}$ and $C_{\text{adv}}$ denote the sets of atomic claims generated by the Skeptic and Advocate, with centroid embeddings $u_{\text{role}} = \frac{1}{|C_{\text{role}}|} \sum_{c \in C_{\text{role}}} \mathbf{e}(c)$ for $\text{role} \in \{\text{sk}, \text{adv}\}$, where $\mathbf{e}(\cdot)$ maps text to dense sentence representations via \texttt{all-MiniLM-L6-v2}. If $\cos(u_{\text{sk}}, u_{\text{adv}}) > 0.85$, role collapse is detected. The guard triggers a contrastive re-run: each role is re-prompted with the opposing agent's claims as explicit negative constraints, forbidding semantic overlap and forcing counter-perspectives.

\subsection{Reflexion and Belief-Delta ($\DB$) Convergence}
Following debate, the Clarification Node generates targeted questions for \texttt{contested} or \texttt{unknown} claims and major Skeptic objections. The Reflexion Node incorporates the resulting evidence into an updated belief state $\mathcal{B}_{t+1}$.

Let $\mathcal{B}_t$ and $\mathcal{B}_{t+1}$ denote the entries in consecutive belief states. Entries are matched by exact equality of their claim-statement strings; let $M_t$ be the set of matched entry pairs $(c_t,c_{t+1})$. A matched entry is considered modified if its categorical status changes or if its confidence shifts by $|\Delta\text{conf}| > 0.10$. Let $m_t$ be the number of modified pairs in $M_t$. Added and dropped entries do not count in $m_t$, although they affect the denominator through the state sizes. We formalize the belief delta as:
\begin{equation}
  \DB_t =
  \begin{cases}
    0, & |\mathcal{B}_t| = |\mathcal{B}_{t+1}| = 0, \\
    \dfrac{m_t}{\max(|\mathcal{B}_t|, |\mathcal{B}_{t+1}|)}, & \text{otherwise}.
  \end{cases}
\end{equation}
$\DB_t$ is evaluated immediately after the Reflexion update. If $\DB_t \geq \tau$ (default $\tau = 0.10$), the revised belief state is routed back to the Asymmetric Pressure Tester for another round of debate. The system halts when $\DB_t < \tau$. Under this matching rule, the value measures the proportion of the larger belief state represented by matched entries with a status or confidence change; it does not by itself establish epistemic stability. Because added and dropped entries do not enter the numerator, and paraphrased statements are treated as unmatched, this rule can understate change when claims are introduced, removed, or reworded.

\subsection{Source-Grounded Evidence Auditing}
Upon convergence, all \texttt{supported} evidence links are audited through two external pathways: (1) an independent NLI cross-encoder (\texttt{cross-encoder/nli-deberta-v3-base}) computes $P(\text{entailment} \mid \texttt{source\_snippet}, \texttt{claim\_text})$; and (2) embedding cosine similarity (\texttt{all-MiniLM-L6-v2}) audits topical match, flagging links with similarity $<0.50$ as embedding-flagged ungrounded. This similarity threshold is a topical-grounding proxy and does not by itself establish that a citation is fabricated.

\section{\uppercase{Experimental Setup}}
\label{sec:setup}

\paragraph{Hypothesis Benchmark.}
We evaluate across 40 research hypotheses split equally between two domains: \textbf{LLM Reasoning \& Self-Correction} (20 hypotheses covering chain-of-thought faithfulness, self-correction limits, and sycophancy) and \textbf{Multi-Agent Coordination} (20 hypotheses covering agent debate, role diversity, memory, and LLM-as-judge biases). Hypotheses were chosen to represent active scientific debates with non-trivial literature evidence.

\paragraph{Inference Backbone.}
All agent roles are powered by \texttt{deepseek-r1:32b} executed locally via Ollama with full chain-of-thought reasoning enabled. Sampling temperature is fixed at 0.0 for reproducibility.

\paragraph{Literature-Grounded Baselines.}
We benchmark \sys{} against four baselines, totaling 200 experimental executions (40 hypotheses $\times$ 5 configurations):
\begin{itemize}
  \item \textbf{Full System (Ours)}: Asymmetric 3-role debate with artifact-grounded $\DB$ stopping.
  \item \textbf{Multi-Agent Debate (MAD)}~\citep{liang2023,du2023}: Symmetric multi-agent debate where Skeptic and Advocate share identical prompt structures.
  \item \textbf{Self-Refine Critic}~\citep{madaan2023}: Monolithic single-model critique in a single pass.
  \item \textbf{Fixed-Step ($N{=}1$)}: Static stopping after exactly 1 iteration.
  \item \textbf{Fixed-Step ($N{=}3$)}: Static stopping after exactly 3 iterations.
\end{itemize}

\paragraph{Reflexion Clarification Protocol.}
In an interactive deployment, \sys{} queries the researcher with targeted clarifying questions from the Reflexion Node. In this autonomous benchmark, clarification questions received a neutral synthetic response (\emph{``I don't have additional context''}) to evaluate convergence behavior without human input. This isolates the system's ability to detect informational exhaustion: a well-calibrated verifier should halt when external feedback provides no new signal rather than querying repeatedly.

\section{\uppercase{Results}}
\label{sec:results}

\subsection{Experiment 1: Role Asymmetry}
\label{sec:exp1}

\begin{table}[ht]
\caption{Experiment 1: Role Asymmetry ($n{=}40$ per configuration, 120 total runs). Lower role similarity indicates greater perspective divergence. Contested fraction and claim diversity capture downstream epistemic quality.}\label{tab:exp1}
\centering
\resizebox{\columnwidth}{!}{
\begin{tabular}{lcccc}
\toprule
\textbf{Configuration} & \textbf{$n$} & \textbf{Role Sim.\ $\downarrow$} & \textbf{Contested $\uparrow$} & \textbf{Diversity $\uparrow$} \\
\midrule
Full System (Ours) & 40 & \textbf{0.852} $\pm$ 0.009 & \textbf{22.0\%} & \textbf{0.539} \\
MAD~\citep{liang2023} & 40 & 0.918 $\pm$ 0.005 & 15.4\% & 0.467 \\
Self-Refine~\citep{madaan2023} & 40 & --- & 20.1\% & 0.512 \\
\bottomrule
\end{tabular}
}
\\[2pt]{\scriptsize Role similarity computed across runs with debate ($n=39$ for Full System and MAD). Welch's $t$-test between Full System and MAD: $t = -6.22, p = 5.82 \times 10^{-8}$; Mann-Whitney $U = 241.0, p = 1.07 \times 10^{-7}$. Single Critic evaluation is monolithic (cross-role divergence is not applicable).}
\end{table}

As shown in Table~\ref{tab:exp1}, \sys{} achieves a mean role-embedding similarity of 0.852 vs.\ 0.918 for MAD ($t = -6.22, p = 5.82 \times 10^{-8}$; $U = 241.0, p = 1.07 \times 10^{-7}$). This divergence translates directly to a higher contested claim fraction (22.0\% vs.\ 15.4\%) and greater claim diversity (0.539 vs.\ 0.467), confirming that asymmetric prompting surfaces a broader evidentiary record.

\begin{figure*}[!t]
  \centering
  \begin{subfigure}[t]{0.4\textwidth}
    \includegraphics[width=\textwidth]{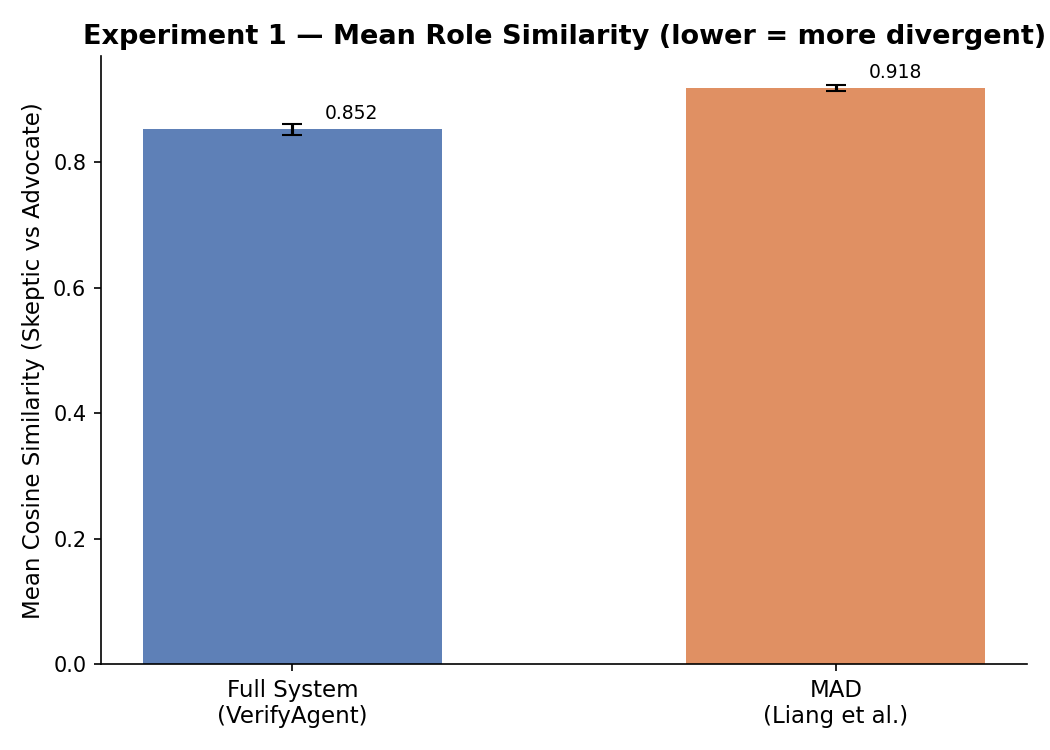}
    \caption{Role-Embedding Similarity (lower = more divergent)}
  \end{subfigure}
  \hfill
  \begin{subfigure}[t]{0.4\textwidth}
    \includegraphics[width=\textwidth]{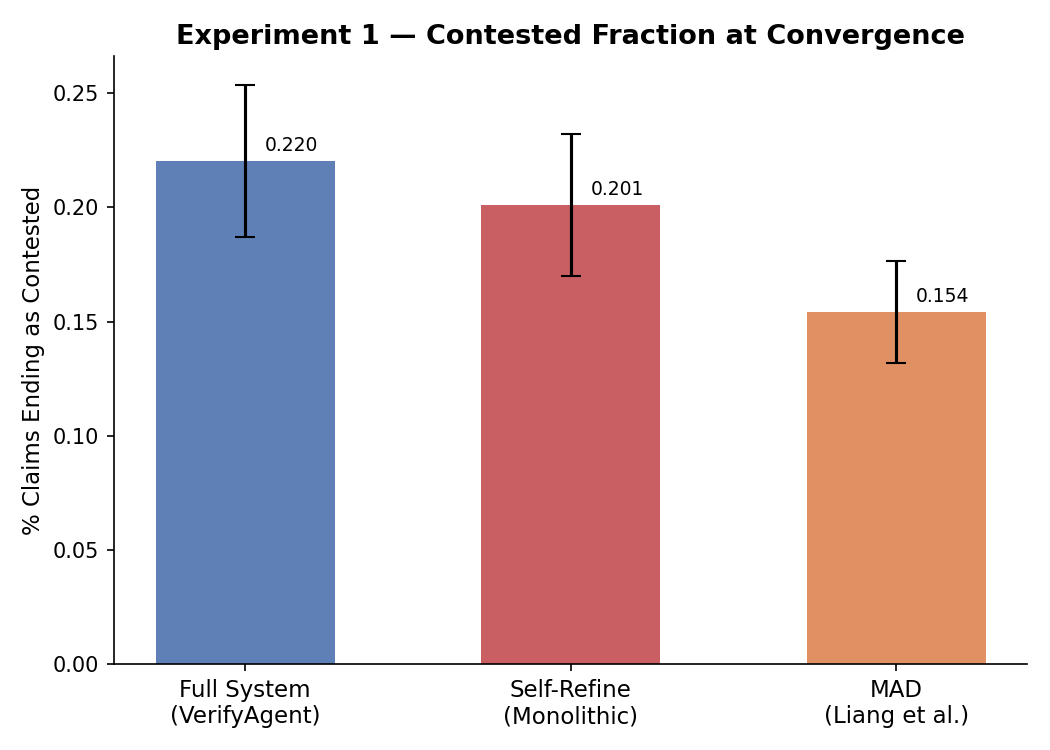}
    \caption{Contested Claim Fraction at Convergence}
  \end{subfigure}
  \\[0.2em]
  \begin{subfigure}[t]{0.4\textwidth}
    \includegraphics[width=\textwidth]{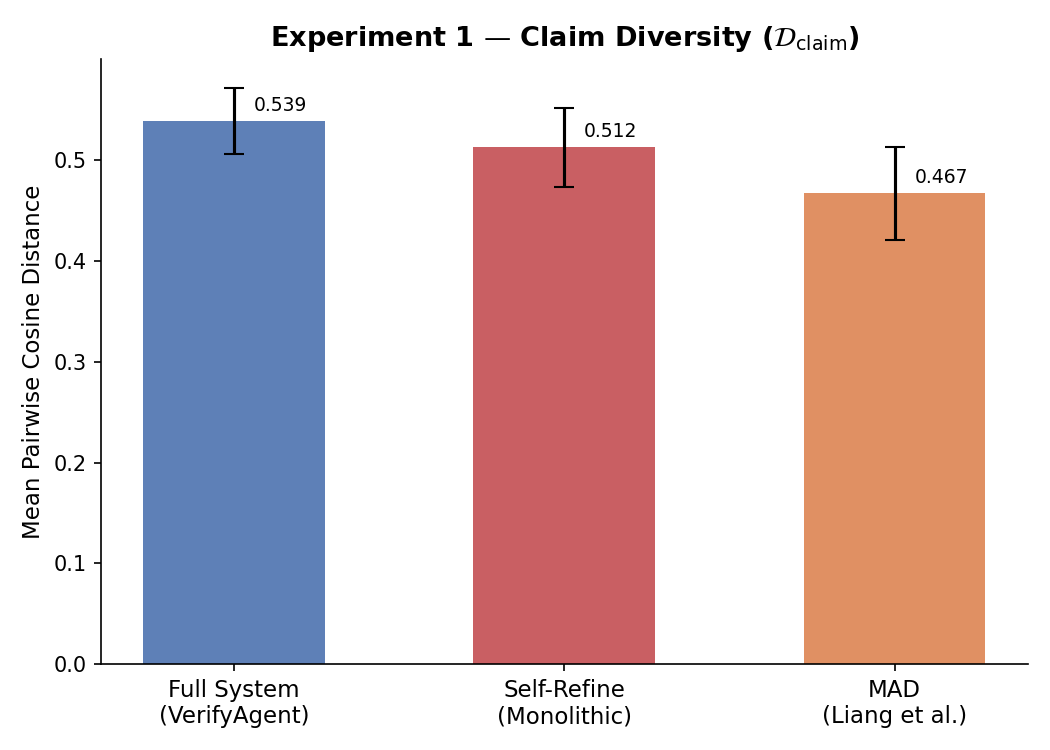}
    \caption{Mean Pairwise Claim Diversity ($\mathcal{D}_{\text{claim}}$)}
  \end{subfigure}
  \hfill
  \begin{subfigure}[t]{0.4\textwidth}
    \includegraphics[width=\textwidth]{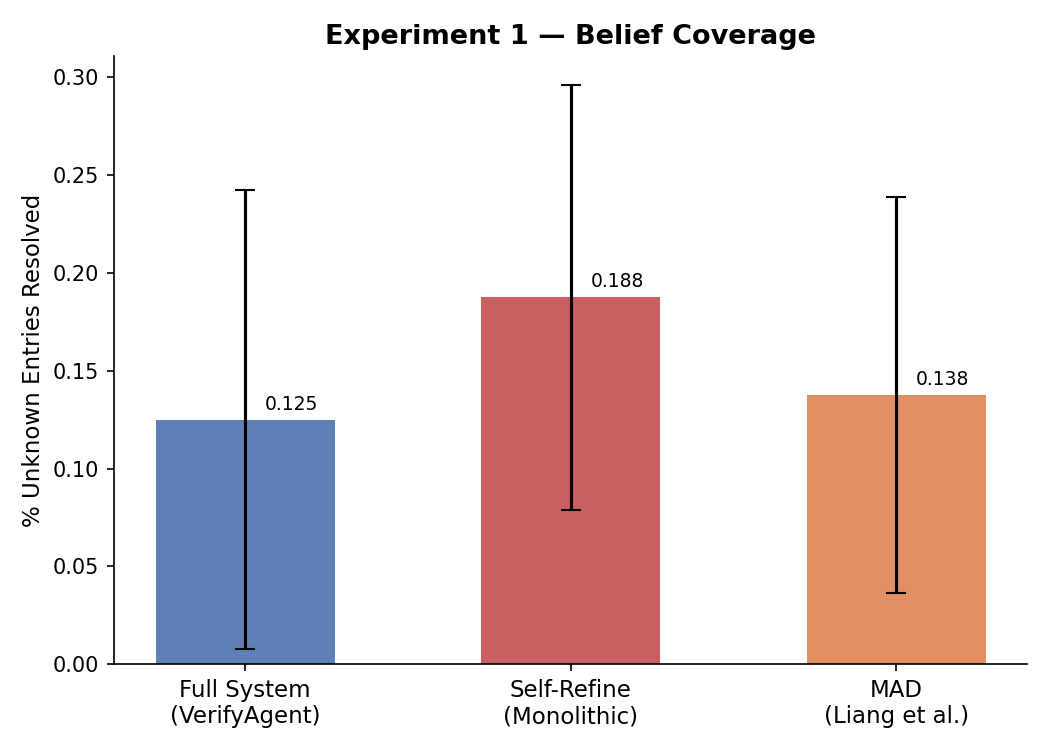}
    \caption{Belief Coverage (resolved unknown claims)}
  \end{subfigure}
  \caption{Experiment 1: Role Asymmetry and Perspective Diversity metrics across 120 experimental runs (mean $\pm$ SEM).}
  \label{fig:exp1}
\end{figure*}

\subsection{Experiment 2: Stopping Calibration}
\label{sec:exp2}

\begin{table}[ht]
\caption{Experiment 2: Stopping Calibration ($n{=}40$ per configuration, 120 total runs). Premature stopping is defined as $>30\%$ claims remaining unresolved at halt.}\label{tab:exp2}
\centering
\resizebox{\columnwidth}{!}{
\begin{tabular}{lcccc}
\toprule
\textbf{Configuration} & \textbf{Premature $\downarrow$} & \textbf{Over-Iter $\downarrow$} & \textbf{LLM Calls $\downarrow$} & \textbf{Iters} \\
\midrule
Full System ($\DB$) & \textbf{77.5\%} & \textbf{0.0\%} & \textbf{8.28} $\pm$ 0.31 & 1.05 $\pm$ 0.05 \\
Fixed-Step ($N{=}1$) & 87.5\% & 0.0\% & 8.00 $\pm$ 0.00 & 1.00 $\pm$ 0.00 \\
Fixed-Step ($N{=}3$) & 85.0\% & 62.5\% & 19.52 $\pm$ 0.48 & 2.92 $\pm$ 0.08 \\
\bottomrule
\end{tabular}
}
\end{table}

Table~\ref{tab:exp2} shows that $\DB$ completely eliminates over-iteration (0.0\%), while Fixed-Step ($N{=}3$) wastes 62.5\% of its budget on already-stabilized belief states, consuming 2.4$\times$ more LLM calls. In 36 of 40 runs, $\DB = 0.0 < 0.10$ after round 1 because the synthetic clarification response carried no new signal, triggering an immediate halt. Fixed-Step ($N{=}3$), unaware of the artifact state, re-ran pressure-testing against the same evidence regardless.

\begin{figure*}[!t]
  \centering
  \begin{subfigure}[t]{0.44\textwidth}
    \includegraphics[width=\textwidth]{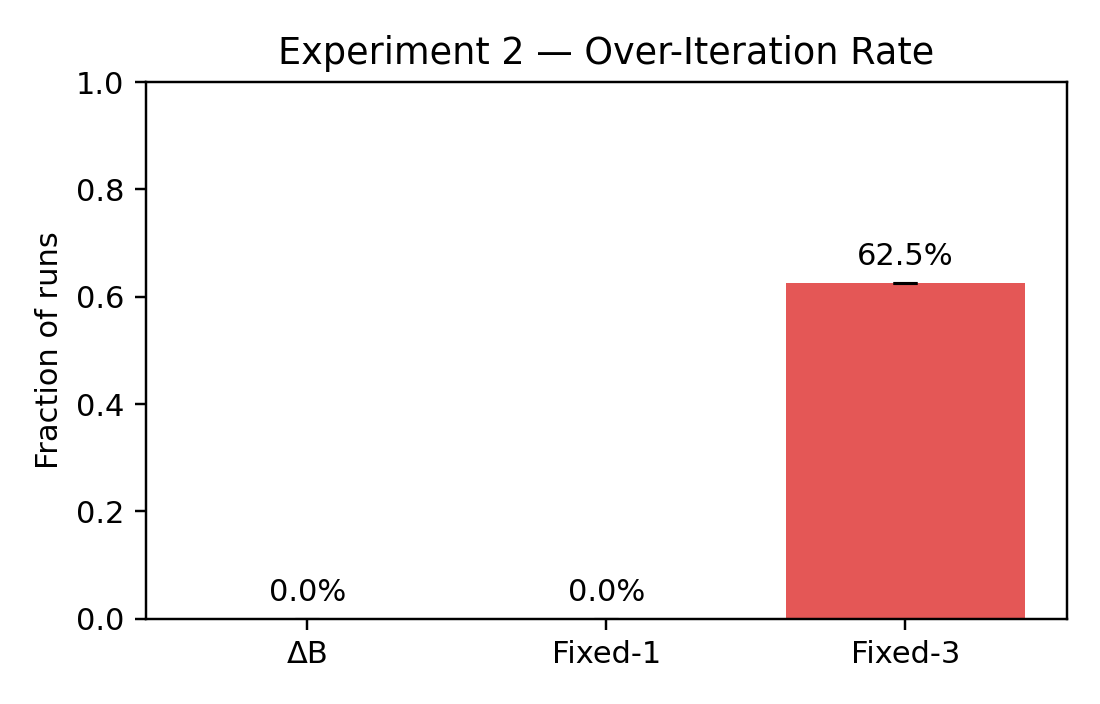}
    \caption{Over-iteration rate from Table~\ref{tab:exp2}}
  \end{subfigure}
  \hfill
  \begin{subfigure}[t]{0.44\textwidth}
    \includegraphics[width=\textwidth]{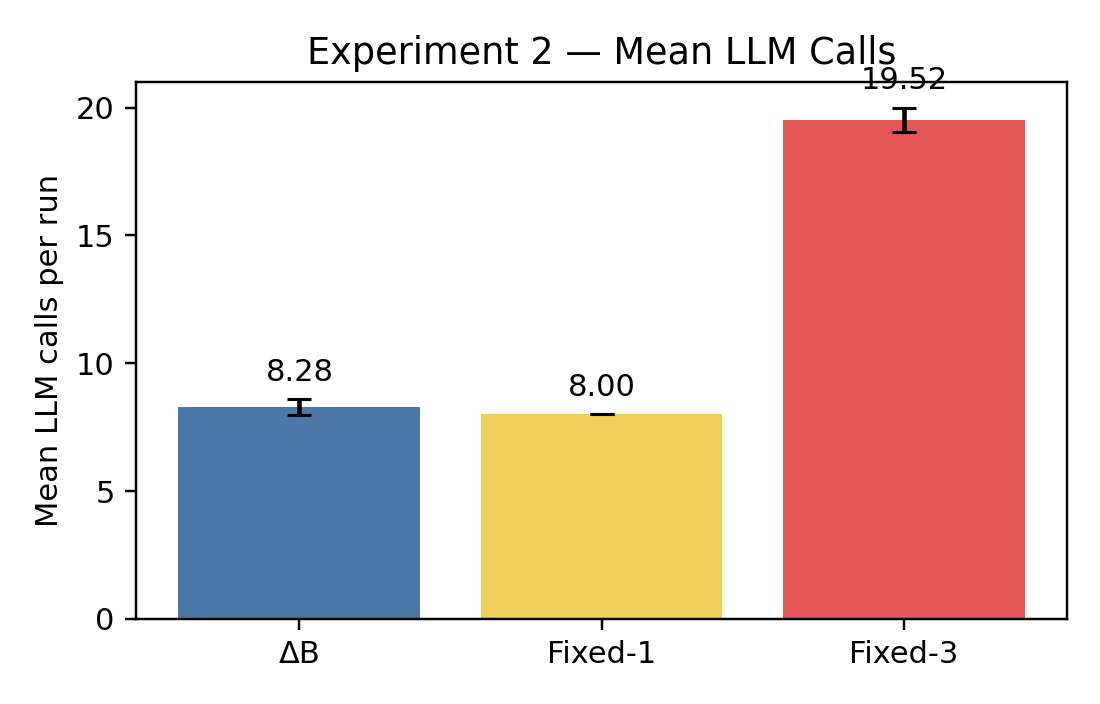}
    \caption{Mean LLM calls per run from Table~\ref{tab:exp2}}
  \end{subfigure}
  \\[0.2em]
  \begin{subfigure}[t]{1.0\textwidth}
    \includegraphics[width=0.96\textwidth]{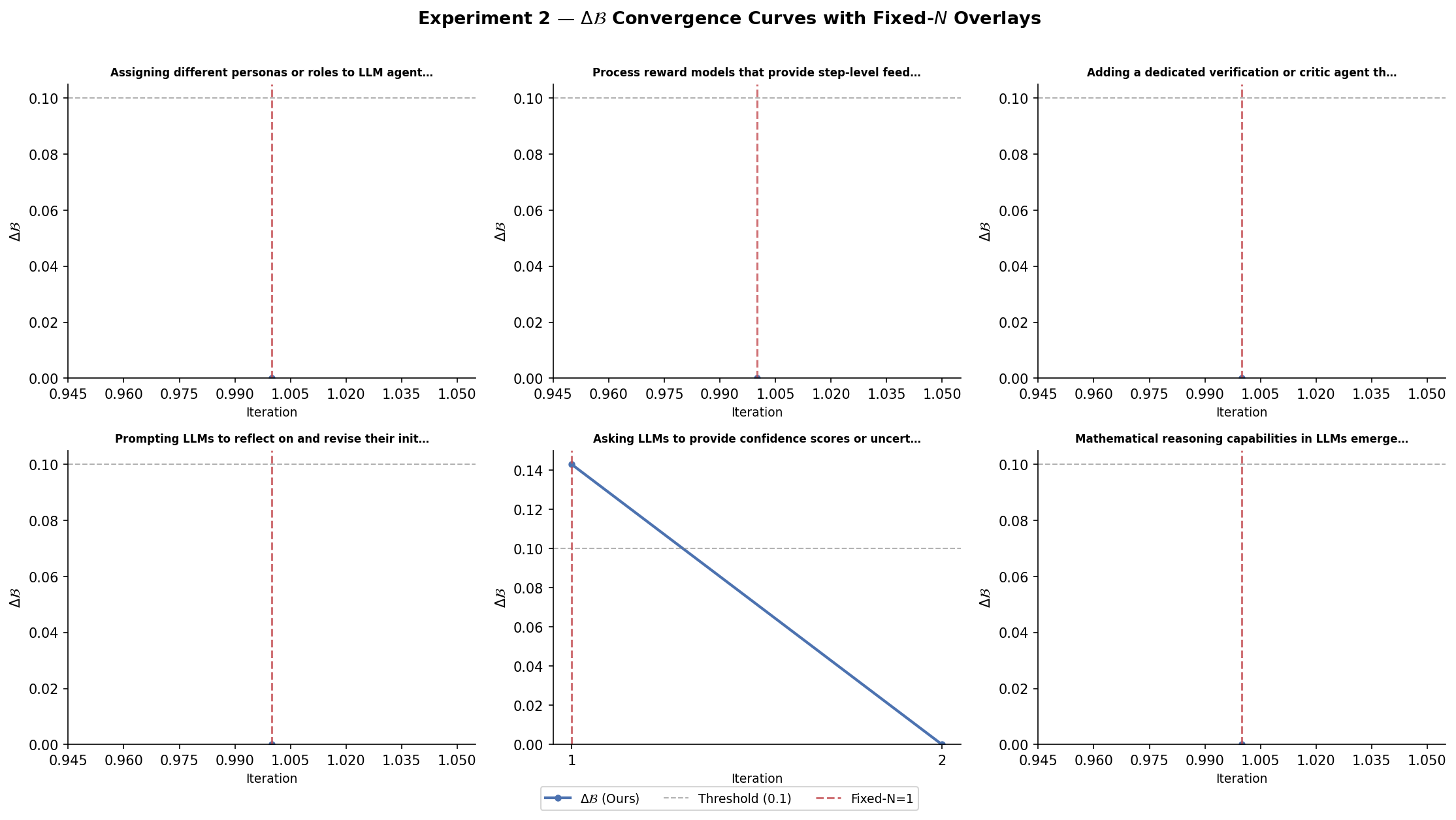}
    \caption{$\DB$ Convergence Dynamics across Representative Hypotheses (dashed lines show static stopping)}
  \end{subfigure}
  \caption{Experiment 2: Stopping calibration and convergence dynamics across 120 experimental runs.}
  \label{fig:exp2}
\end{figure*}

\subsection{Experiment 3: Evidence Grounding}
\label{sec:exp3}

\begin{table}[ht]
\caption{Experiment 3: Evidence Grounding evaluated over runs producing supported claims.}\label{tab:exp3}
\centering
\resizebox{\columnwidth}{!}{
\begin{tabular}{lcccc}
\toprule
\textbf{Configuration} & \textbf{$n$} & \textbf{Entailment $\uparrow$} & \textbf{Hallucination $\downarrow$} & \textbf{Empty $\downarrow$} \\
\midrule
Full System (Ours) & 38 & 0.338 $\pm$ 0.03 & 21.7\% $\pm$ 5.4\% & 2.6\% \\
Self-Refine Critic & 34 & 0.340 $\pm$ 0.04 & 25.9\% $\pm$ 6.3\% & \textbf{0.0\%} \\
\bottomrule
\end{tabular}
}
\\[2pt]{\scriptsize Entailment difference: $t = -0.05, p = 0.96$. Hallucination difference: $t = -0.509, p = 0.613$; Mann-Whitney $U = 615.5, p = 0.702$.}
\end{table}

Table~\ref{tab:exp3} shows nearly identical entailment scores (0.338 vs.\ 0.340, $p = 0.96$), consistent with LLM-extracted claims synthesizing high-level conclusions rather than paraphrasing short snippets. The full system achieves a directional hallucination reduction from 25.9\% to 21.7\% ($p = 0.613$); while not significant at this scale, the structural advantage of \sys{} is that every assertion is anchored to an explicit, auditable 120-character snippet. The 2.6\% empty-snippet rate reflects validation retries exhausting the context limit, which unconstrained single-pass generation avoids.

\begin{figure*}[!t]
  \centering
  \begin{subfigure}[t]{0.44\textwidth}
    \includegraphics[width=\textwidth]{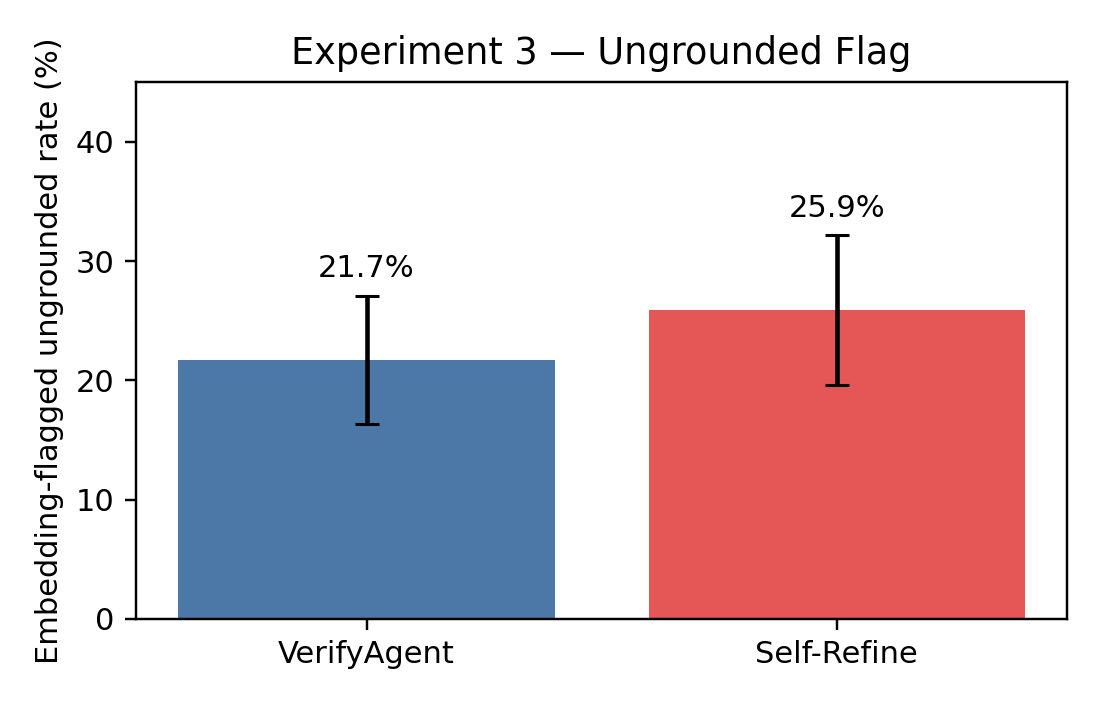}
    \caption{Hallucination Rate}
  \end{subfigure}
  \hfill
  \begin{subfigure}[t]{0.44\textwidth}
    \includegraphics[width=\textwidth]{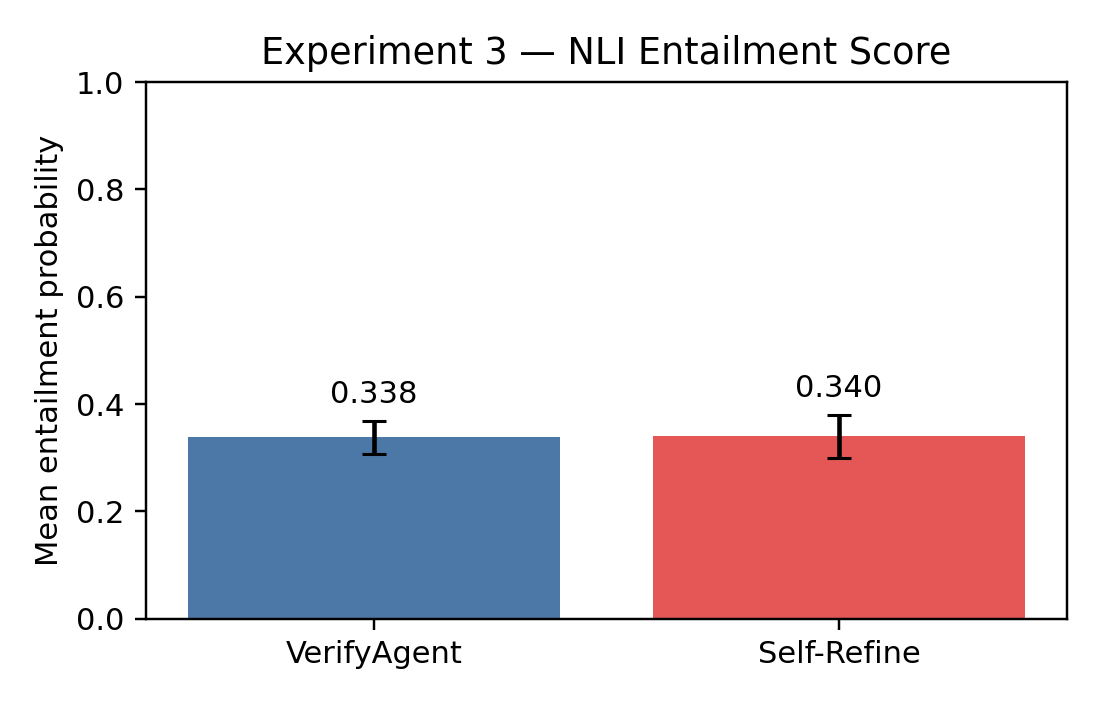}
    \caption{Mean NLI Score}
  \end{subfigure}
  \caption{Experiment 3: Evidence Grounding and Hallucination Reduction audited by \texttt{cross-encoder/nli-deberta-v3-base}.}
  \label{fig:exp3}
\end{figure*}

\section{\uppercase{Discussion}}
\label{sec:discussion}

\paragraph{Epistemic Stagnation and Circuit Breaking.}
Belief coverage reaches only 12.5\%--18.8\% across all configurations. This is expected: abstracts present conditional findings rather than binary confirmations, \texttt{deepseek-r1:32b} is conservative under conflicting evidence, and several hypotheses address controversies lacking scientific consensus. The $\DB$ criterion proves its value here: rather than looping blindly as Fixed-$3$ does (62.5\% over-iteration), it detects state stabilization and halts immediately.

\paragraph{Interpreting Structural Asymmetry.}
The role similarity divergence (0.852 vs.\ 0.918, $p = 5.82 \times 10^{-8}$) confirms that prompt-level differentiation prevents symmetric collapse. In the single run where similarity could not be computed, retrieved abstracts were unanimously critical, leading the Advocate to correctly propose zero affirmative claims, demonstrating that agents do not fabricate support when the literature provides none.

\paragraph{Limitations.}
Our evaluation has three main limitations: (1) all experiments used a local model (\texttt{deepseek-r1:32b}); frontier proprietary models with larger context windows may achieve higher belief resolution; (2) the benchmark covers 40 hypotheses, and expanded benchmarks with crowd-sourced ground-truth adjudications would enable end-to-end precision-recall evaluation; and (3) cross-encoder NLI models trained on general-domain data occasionally assign conservative entailment scores to valid scientific paraphrases.

\section{\uppercase{Conclusion}}
\label{sec:conclusion}

\noindent We presented \sys{}, a multi-agent research verification framework designed to address self-critique bias and miscalibrated stopping. By enforcing an asymmetric Advocate/Skeptic/Arbitrator architecture, \sys{} achieves statistically significant role divergence ($p = 5.82 \times 10^{-8}$) and surfaces contested claims at a higher rate (22.0\% vs.\ 15.4\%) than standard Multi-Agent Debate. The artifact-grounded $\DB$ convergence criterion eliminates over-iteration and reduces LLM calls by 2.4$\times$. Verbatim source snippet grounding, audited by cross-encoder NLI, provides an auditable evidence trail suitable for automated scientific literature evaluation. All code, prompts, and experimental trajectories are included in the supplementary material and will be publicly released upon publication.

\newpage

\bibliographystyle{apalike}
\bibliography{references}

\end{document}